\documentclass[10pt,twocolumn,letterpaper]{article}

\usepackage{cvpr}
\usepackage{times}
\usepackage{epsfig}
\usepackage{graphicx}
\usepackage{amsmath}
\usepackage{amssymb}
\usepackage{mathtools, cuted}
\usepackage{afterpage}
\usepackage{algorithm, algorithmic}

\newcommand\blfootnote[1]{%
  \begingroup
  \renewcommand\thefootnote{}\footnote{#1}%
  \addtocounter{footnote}{-1}%
  \endgroup
}

\usepackage[pagebackref=true,breaklinks=true,letterpaper=true,colorlinks,bookmarks=false]{hyperref}

\cvprfinalcopy % *** Uncomment this line for the final submission

\def\cvprPaperID{5477} % *** Enter the CVPR Paper ID here

\ifcvprfinal\fi
\begin{document}

%%%%%%%%% TITLE
\title{NestedVAE: Isolating Common Factors via Weak Supervision.}

\author{Matthew J. Vowels\\
{\tt\small m.j.vowels@surrey.ac.uk}
\and
Necati Cihan Camgoz\\
{\tt\small n.camgoz@surrey.ac.uk}
\and
Richard Bowden \\
{\tt\small r.bowden@surrey.ac.uk}
\and
Centre for Vision, Speech and Signal Processing \\
University of Surrey\\
Guildford, UK\\
}

\maketitle
%\thispagestyle{empty}

%%%%%%%%% ABSTRACT
\begin{abstract}
Fair and unbiased machine learning is an important and active field of research, as decision processes are increasingly driven by models that learn from data. Unfortunately, any biases present in the data may be learned by the model, thereby inappropriately transferring that bias into the decision making process. We identify the connection between the task of bias reduction and that of isolating factors common between domains whilst encouraging domain specific invariance. To isolate the common factors we combine the theory of deep latent variable models with information bottleneck theory for scenarios whereby data may be naturally paired across domains and no additional supervision is required. The result is the Nested Variational AutoEncoder (NestedVAE). Two outer VAEs with shared weights attempt to reconstruct the input and infer a latent space, whilst a nested VAE attempts to reconstruct the latent representation of one image, from the latent representation of its paired image. In so doing, the nested VAE isolates the common latent factors/causes and becomes invariant to unwanted factors that are not shared between paired images.  We also propose a new metric to provide a balanced method of evaluating consistency and classifier performance across domains which we refer to as the Adjusted Parity metric. An evaluation of NestedVAE on both domain and attribute invariance, change detection, and learning common factors for the prediction of biological sex demonstrates that NestedVAE significantly outperforms alternative methods.
\end{abstract}

\vspace{-2em}

\section{Introduction}
\blfootnote{Paper accepted to CVPR 2020.}
One of the goals of representation learning is to achieve an embedding that informatively captures the underlying factors of variation in data \cite{bengio1}. However, many techniques for learning such embeddings have been found to also learn unwanted or confounding factors, irrelevant or detrimental to the intended task(s) \cite{DIVA}. Such factors can include distribution specific bias, which impairs the generalizability of a model across empirical samples or in the face of distributional shift \cite{bousmalis2016, DIVA, shankar2018, bengio2019}, or bias associated with culturally sensitive or legally protected characteristics such as race, age, gender or sex \cite{locatello2019fairness, cao2019, liu2019, howard2018, rose2010, louizos2017, moyer1, gendershades}. 

 \begin{figure}[]
\centering
\includegraphics[width=0.8\linewidth]{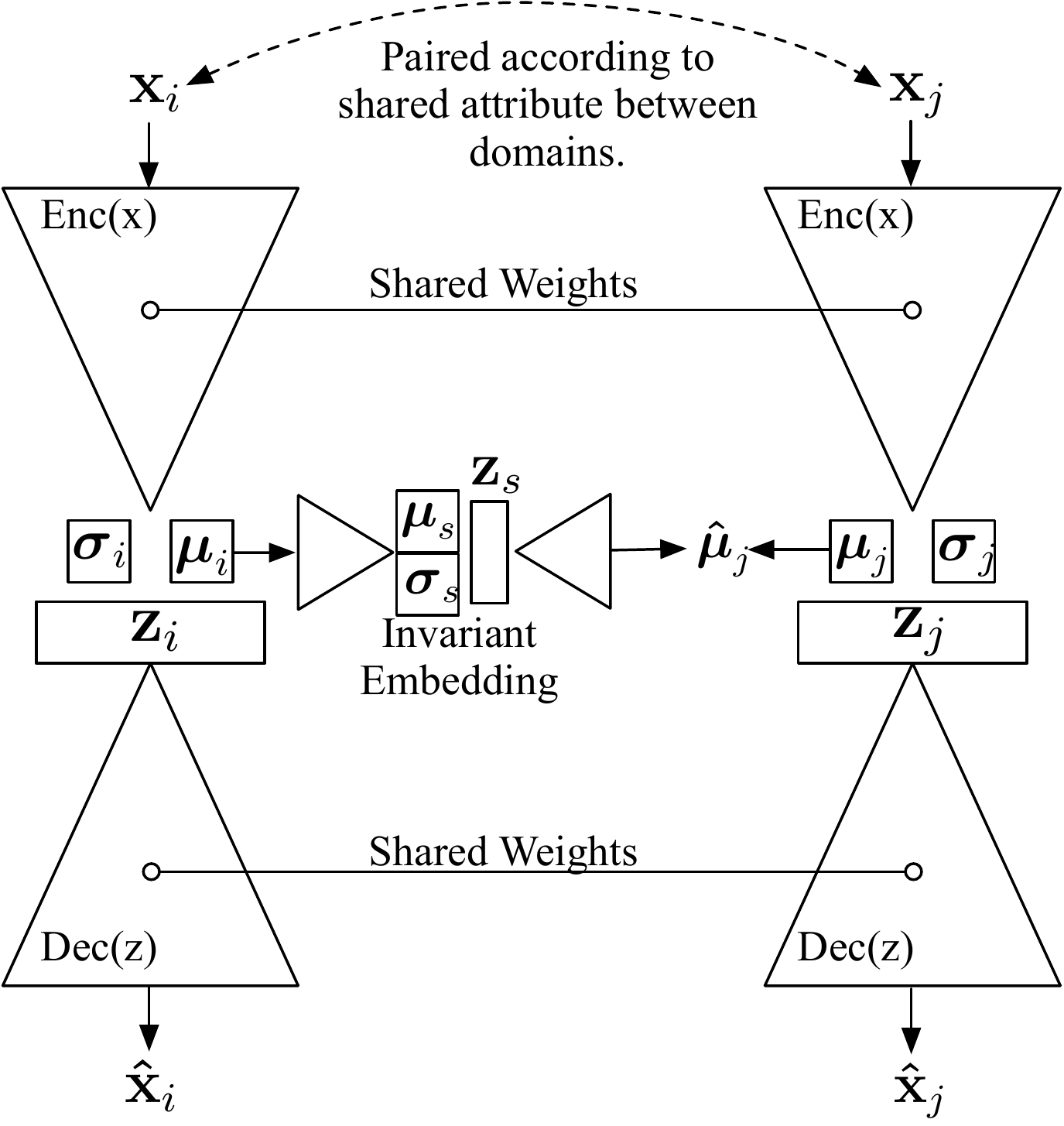}
\caption{Top-level architecture of NestedVAE. Images (or alternative data modality) are paired according to shared attributes or domains. Latent representations $\mathbf{z}_i$ and $\mathbf{z}_j$ for images $\mathbf{x}_i$ and $\mathbf{x}_j$ are derived and fed to a secondary `nested' VAE. Using the principles from Information Bottleneck theory, a sufficient and minimal representation $\mathbf{z}_s$ for $\mathbf{z}_j$ may be derived from $\mathbf{z}_i$ and vice versa. $\mathbf{z}_s$ may therefore be interpreted as representing the common factors, or common causes for the two images. Sufficiency indicates it contains the information common to both, and minimality indicates that it is invariant to information specific to each. }
\label{fig:nested}
\end{figure}

Indeed, the prevalence of reports of systemic bias arising from the use of machine learning algorithms is increasing \cite{holstein2019, nabi2019, slack2019}. Furthermore, conceptually distinct factors, such as object type and pose, may be entangled in the embedding, despite a prior expectation that they ought to be factorized. Learning models that solve these problems is therefore important from a number of converging engineering and societal perspectives \cite{locatello2019fairness}. In terms of engineering, we may wish for our models to be informative, to be invariant to nuisance factors, to perform well and generalize across domains, and to disentangle independent factors of variation. From a societal perspective, we may wish to achieve statistical and demographic parity such that our models do not reflect or amplify any unfairness present in our data or in society itself \cite{moyer1,zemel2013, holstein2019, nabi2019}. 

\begin{figure}
\centering
\includegraphics[width=0.8\linewidth]{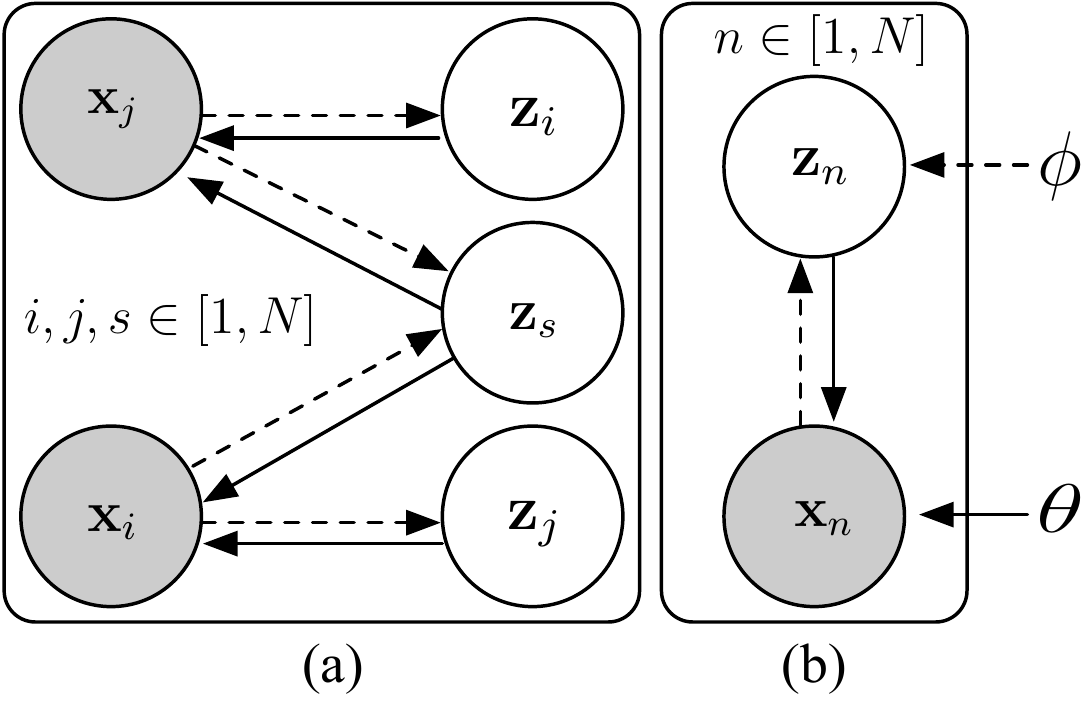}
\caption{Probabilistic Graphic Model for (a) inferring the common factors $\mathbf{z_s}$ from pairs of images $\mathbf{x}_i$ and $\mathbf{x}_j$ and (b) the inference and generative processes of a VAE. Dotted lines indicate inference and solid lines indicate generation. $\phi$ and $\theta$ are the VAE encoder (inference) and decoder (generation) parameters respectively. }
\label{fig:PGM}
\end{figure}

Success at these overlapping tasks has implications for a range of more specific downstream tasks including attribute transfer \cite{szabo1, szabo2018, higgins, jha2018, zheng2018}, person re-identification \cite{JADE, ganin1}, change detection \cite{hamaguchi} adversarial robustness \cite{puvae}, and machine learning based decision processes \cite{locatello2019fairness, barocas2019, mehrabi2019}.

The contributions of this work are as follows:

\begin{itemize}
    \item A unified interpretation of prior work on bias, disentanglement, fairness, domain/attribute invariance, and common causes.
    \item A novel deep latent variable model called the Nested Variational Autoencoder (NestedVAE) that combines deep, amortized variational inference \cite{kingma} and Information Bottleneck (IB) theory \cite{tishby2000, tishby2015}.
    \item A demonstration that NestedVAE achieves significant improvements in classification and regression performance by learning the factors that are common between domains.
    \item A novel metric for evaluating regression and classification parity across domains, referred to as the Adjusted Parity Metric, that accounts for both overall predictive performance as well the variation in performance across domains.
\end{itemize}

\section{Formulation}
\subsection{Problem Formulation}
We consider the problem of encoding an informative, latent representation $\mathbf{z} \sim p(\mathbf{z})$ from observation $\mathbf{x} \sim p(\mathbf{x}|\mathbf{z},\mathbf{c})$  such that $\mathbf{z}$ is invariant to some irrelevant/nuisance/confounding covariate $\mathbf{c}$ \cite{louizos}. From a statistical parity perspective, we wish to be able to use the latent representation for some arbitrary downstream prediction of label $\mathbf{y}$ such that $p(\hat{y} = y | \mathbf{c},\mathbf{z} ) = p(\hat{y} = y | \mathbf{z})$ $ \forall y, \mathbf{c}, \mathbf{z}$ \cite{locatello}. We therefore wish for $\mathbf{z} \perp \!\!\! \perp  \mathbf{c}$ and $y \perp \!\!\! \perp  \mathbf{c}$. From a domain invariance perspective, we wish learning to transfer as much as possible between the different domains, where each domain is associated with its own confounders or covariates. In other words, the latent representation that we learn should be independent of nuisance or confounding factors, thereby also resulting in downstream task performance that is invariant to these factors. Further, the resulting representation will represent the latent factors common to each domain.

For the development of NestedVAE, we consider the incorporation of weak supervision whereby the supervision takes the form of data pairs \cite{shu2019}. Scenarios whereby natural pairings occur or may be straightforwardly derived include: time series data whereby individuals appearing in frames from the same scene vary in terms of pose and expression but maintain identity \cite{denton2017, gabbay2019}; pairings within domains where the domains may be hospitals or patients and the data may be medical images \cite{DIVA}; pairings arising from dyadic interactions such as conversations \cite{RN11}; and pairings of data representing images of objects (e.g. hands from sign language data) from multiple viewpoints. Without loss of generality we primarily consider the application of computer vision with face images, whereby the images are paired according to sex.

For the following formalization we assume two domains, although the model can be extended to include any number of domains for which we can form pairs. The Probabilistic Graphic Model (PGM) corresponding with our world model is depicted in Figure \ref{fig:PGM}a.  We assume that each image $\mathbf{x}_i \in X_1$ and $\mathbf{x}_j \in X_2$ has latent factors/causes $\mathbf{z}_i$ and $\mathbf{z}_j$ specific to the respective domains $X_1$ or $X_2$, as well as shared factors/causes $\mathbf{z}_s$ common to both domains. From the perspective of learning domain invariance, $\mathbf{z}_i$ and $\mathbf{z}_j$ represent confounders $\mathbf{c}_i$ and $\mathbf{c}_j$ respectively, and $X_1$ and $X_2$ represent different domains to which the representation $\mathbf{z}_s$ should be agnostic/invariant. From the perspectives of causal modelling, $\mathbf{z}_i$ and $\mathbf{z}_j$ are domain specific latent causes, and $\mathbf{z}_s$ are common latent causes \cite{lee2019}. This is similar to the confounding additive noise model \cite{janzing2009, lee2019} where $\mathbf{x}_i = f_i(\mathbf{z}_i) + g_i(\mathbf{z}_s) + \boldsymbol{\epsilon}_i$ and $\mathbf{x}_j = f_j(\mathbf{z}_j) + g_j(\mathbf{z}_s) + \boldsymbol{\epsilon}_j$, where $f$ and $g$ are arbitrary functions, and $\boldsymbol{\epsilon}$ is additive noise.

For each pair of images we wish to learn a representation $\mathbf{z_s}$ that represents only the common factors between images in the pair. In order to do so, we leverage the information gain achieved from specific pairings in order to infer $\mathbf{z}_s$ from $\mathbf{z}_i$ and $\mathbf{z}_j$, and take inspiration from the information bottleneck perspective \cite{tishby2000, tishby2015, moyer1, achille2018}. To do so, we model the shared and common factors as a Markov chain: 

\vspace{-1em}

\begin{equation}
\mathbf{z}_i \longrightarrow \mathbf{z}_s \longrightarrow \mathbf{z}_j \mbox{ s.t. } p(\mathbf{z}_j | \mathbf{z}_i, \mathbf{z}_s) = p(\mathbf{z}_j | \mathbf{z}_s)
\label{eq:markov}
\end{equation}

The Data Processing Inequality \cite{cover2006} means that $\mathbf{z}_s$ cannot contain more information about $\mathbf{z}_j$ than $\mathbf{z}_i$ \cite{achille2018}. The information about $\mathbf{z}_j$ in $\mathbf{z}_s$ can therefore only be what is common to both $\mathbf{z}_i$ and $\mathbf{z}_j$. The image pairings are formed as non-ordered combinations such that we pair $\mathbf{x}_i$ with $\mathbf{x}_j$ but also $\mathbf{x}_j$ with $\mathbf{x}_i$. As such, our task becomes that of predicting $\mathbf{z}_j$ from $\mathbf{z}_i$ via $\mathbf{z}_s$. Finally, if we make the (albeit strong) assumption that $\mathbf{z}_i \approx \mathbf{z}_j + \epsilon $, where $\epsilon$ represents random perturbations specific to the respective domain, then we can apply VAEs to the task of learning the minimal and sufficient representation $\mathbf{z}_s$ by seeking to generate $\mathbf{z}_i$ from $\mathbf{z}_j$, and vice versa. Sufficiency describes the Markov chain condition in Eq. \ref{eq:markov} whereby $I(\mathbf{z}_s; \mathbf{z}_j) = I(\mathbf{z}_i; \mathbf{z}_j)$, and minimality describes the fact that there is minimal redundant information content in the representation \cite{achille2018, cover2006}.\footnote{Here, $I(.;.)$ is the Shannon mutual information.} In other words, $\mathbf{z}_s$ only contains the information in $\mathbf{z}_j$ which is also in $\mathbf{z}_i$. \\

\subsection{VAEs}

We now turn our attention to VAEs. For a detailed review of the theory, interested readers are directed to \cite{doersch, kingma, rezende2}. The PGM for the inference and generation (or, equivalently, encoding and decoding) processes of the VAE is shown in Figure \ref{fig:PGM}b. Following the theory for variational inference \cite{bishop} for a distribution of latent variables $\mathbf{z}$, we start by sampling $\mathbf{z} \sim p(\mathbf{z})$ and generate dataset $X$ of images $\mathbf{x} \in \mathbb{R}^N$ with reconstructed/generated distribution $p_\theta(\mathbf{x}|\mathbf{z})$. We may derive an inferred posterior for the conditional latent distribution as $q_\phi(\mathbf{z}|\mathbf{x})$ that approximates the true conditional inference distribution $p_\theta(\mathbf{z}|\mathbf{x})$. Both $q_\phi(\mathbf{z}|\mathbf{x})$ and $p_\theta(\mathbf{x}|\mathbf{z})$ are parameterised by neural network encoder and decoder parameters $\phi$ and $\theta$ respectively \cite{tshannen, doersch,infovae}. The approximating distribution $q$ is chosen to circumvent the intractability of the integral when computing (in order to maximize) the marginal likelihood $p(\mathbf{x}) = \int p_{\theta}(\mathbf{x}|\mathbf{z})p(\mathbf{z}) d\mathbf{z}$ and is introduced according to the identity trick:

\begin{equation}
    \log p(\mathbf{x}) = \log \int p_{\theta}(\mathbf{x}|\mathbf{z})\frac{p(\mathbf{z})}{q_{\phi}(\mathbf{z}|\mathbf{x})}q_{\phi}(\mathbf{z}|\mathbf{x})d\mathbf{z}
\end{equation}
 This may be further manipulated to establish a lower bound on the marginal log likelihood $\log p(\mathbf{x})$:
\begin{equation}
\begin{split}
    \log p_{\theta}(\mathbf{x}) = \\
    \mathbb{E}_{\mathbf{z}\sim q_{\phi}(\mathbf{z}|\mathbf{x})} \left[\log p_{\theta}(\mathbf{x}|\mathbf{z})\right] - \mathrm{KL}\left[q_{\phi}(\mathbf{z}|\mathbf{x})||p(\mathbf{z})\right] +... \\ ... + \mathrm{KL}\left[q_{\phi}(\mathbf{z}|\mathbf{x})||p_{\theta}(\mathbf{z}|\mathbf{x})\right]
    \end{split}
    \label{eq:ELBO}
\end{equation}
The last term on the right hand side of Eq. \ref{eq:ELBO} represents the divergence between our true inference distribution and our choice of approximating distribution, and forms what is known as the `approximation gap' between the true log likelihood, and its estimation \cite{mattei2018}. Once we choose our approximating distribution and optimise it, we are unable to reduce this divergence further. This term is usually omitted such that we are left with what is known as either the Variational Lower Bound (VLB) or the Evidence Lower Bound (ELBO), which serves as a proxy for the log-likelihood. We can then maximize the ELBO as follows \cite{kingma,kumar2}:
\begin{equation}
\begin{split}
 \max _{\theta, \phi}\mathbb{E}_{\mathbf{x}}\left[ \mathcal{L}_{\mathrm{ELBO}}(x)\right] =
\\ \max _{\theta, \phi} \mathbb{E}_{\mathbf{x}}\left[\mathbb{E}_{\mathbf{z} \sim q_{\phi}(\mathbf{z} | \mathbf{x})}\left[\log p_{\theta}(\mathbf{x} | \mathbf{z})\right]- \beta \mathrm{KL}\left(q_{\phi}(\mathbf{z} | \mathbf{x}) \| p(\mathbf{z})\right)\right]
\end{split}
\label{eq:ELBO2}
\end{equation}
The first term on the RHS of Eq. \ref{eq:ELBO2} encourages reconstruction accuracy, and the Kullback-Liebler divergence term (weighted by parameter $\beta$ \cite{higgins}) acts as a prior regularizer, penalising approximations for $q_\phi(\mathbf{z}|\mathbf{x})$ that do not resemble the prior. The objective is therefore to maximise the lower bound to the marginal log-likelihood of $\mathbf{x}$ over the latent distribution $\mathbf{z}$ \cite{higgins}, which is assumed to be Gaussian with identity covariance $\mathbf{z} \sim \mathcal{N}(0, \mathbf{I})$. If sample quality is not of primary concern, there is some incentive to weaken the decoder capacity in order to maintain pressure to encode useful information in the latent space (i.e. increase $I(x;z)$) and to prevent decoupling of the decoder from the encoder \cite{lucas2018}. The assumption of Gaussianity means that Eq. \ref{eq:ELBO2} may be written using an analytical reduction of the KL divergence term \cite{kumar2}:
\begin{equation}
\label{eq:ELBO3}
\begin{split}
\max_{\theta, \phi}\mathbb{E}_{\mathbf{x}} 
    \left[ 
        \mathcal{L}_{\mathrm{ELBO}}(x) 
    \right] 
    = 
    \max_{\theta, \phi} \mathbb{E}_{\mathbf{x}}
    \biggl[ 
        \mathbb{E}_{\mathbf{z} \sim q_{\phi}(\mathbf{z} | \mathbf{x})}
        \left[ 
            \log p_{\theta}(\mathbf{x} | \mathbf{z}) 
        \right] - \\ 
        \frac{\beta}{2}
        \bigl(
            \sum_{i}
            \left(
                \left[
                    \Sigma_{\phi}(\mathbf{x})
                \right]_{ii} - \ln 
                \left[ 
                    \boldsymbol{\Sigma}_{\phi}(\mathbf{x})
                \right]_{ii}
            \right) + 
            \left\|
                \boldsymbol{\mu}_{\phi}(\mathbf{x})
            \right\|_{2}^{2}
        \bigr)
    \biggr] 
\end{split}
\end{equation}

In Eq. \ref{eq:ELBO3} the $\left[\Sigma_{\phi}(\mathbf{x})\right]_{i i}$ indicates the diagonal covariance, and $\boldsymbol{\mu}_{\phi}(\mathbf{x})$ is the mean. Both the mean and covariance are learned by the network encoder and parameterize a multivariate Gaussian that forms the inferred latent distribution $q_\phi(\mathbf{z}|\mathbf{x})$. The decoder network samples from  $\mathbf{z} \sim q_\phi(\mathbf{z}|\mathbf{x})$ using the reparameterization trick \cite{doersch} such that $\mathbf{z} = \mu_\phi(\mathbf{x}) + \epsilon\sqrt{\Sigma_\phi(\mathbf{x})}$ where we redefine $\epsilon$ as $\epsilon = \mathcal{N}(0, \mathbf{I})$. One interpretation of disentanglement posits that it is achieved if $q_\phi (\mathbf{z}) = \int{ q_\phi(\mathbf{z}|\mathbf{x})p(\mathbf{x})d\mathbf{x}} = \prod_i q_i(\mathbf{z}_i)$ \cite{kumar2}.

Applying VAEs to our task: we can learn the latent factors $\mathbf{z}_i$ and $\mathbf{z}_j$ for images $\mathbf{x}_i \sim X_1$ and $\mathbf{x}_j \sim X_2$ respectively. The following section describes the means to utilise these embeddings to learn the shared factors $\mathbf{z}_s$.

\subsection{Combining VAEs and Information Bottleneck}

VAEs are closely related to information bottleneck theory through the Information Bottleneck Lagrangian \cite{tshannen, alemi, alemi2017, moyer1}:

\begin{equation}
    \mathcal{L}(p(\mathbf{z}|\mathbf{x})) = H(\mathbf{y}|\mathbf{z}) + \beta I(\mathbf{z};\mathbf{x})
\end{equation}
Notice that $H$, the Shannon entropy of the conditional distribution, is equivalent to the cross-entropy reconstruction term in Eq. \ref{eq:ELBO2}, except that in VAEs the target $\mathbf{y}$ is $\mathbf{x}$ and the network generates a reconstruction $\hat{\mathbf{x}} \sim p(\hat{\mathbf{x}}|\mathbf{z})$. Further, notice that $I(\mathbf{z};\mathbf{x}) = \mathbb{E}_{\mathbf{x}}\mathrm{KL}\left[q_{\phi}(\mathbf{z}|\mathbf{x})||p(\mathbf{z})\right] $ which is the prior regularizer in Eq. \ref{eq:ELBO2}. Finally the $\beta$ term is proposed to be learned via Lagrangian optimization \cite{achille2018} although, for VAEs, it may also be annealed during training \cite{disentanglement} or evaluated as a hyperparameter \cite{higgins}.

Making the assumption that $\mathbf{z}_i \approx \mathbf{z}_j + \epsilon $, we can reapply the VAE model to this problem. As such, we apply an `outer' VAE to the problem of learning $\mathbf{z}_i$ and $\mathbf{z}_j$ and a `nested' VAE to the problem of learning the common factors $\mathbf{z}_s$. The full loss function over $\mathbf{x}_i \sim X_1$ and $\mathbf{x}_j \sim X_2$ is simply a combination of the outer and nested VAE objectives for each image in a pair, and is presented in Eq. \ref{eq:full}. Here, $\phi_1$, $\theta_1$,  and $\phi_2$, $\theta_2$ are the encoder and decoder parameters for the `outer' and `nested' VAEs respectively. We have assumed the same prior distribution $p(\mathbf{z})$ and the same approximating distribution family $q$ for both outer and nested VAEs.

\vspace{-2em}

\begin{equation}
\begin{split}
 \max _{\theta_1, \phi_1,\theta_2, \phi_2}\mathbb{E}_{\mathbf{x}_i\sim X_1,\mathbf{x}_j\sim X_2,  }\left[ \mathcal{L}_{\mathrm{Nested}}\right] = \\
\max _{\theta_1, \phi_1,\theta_2, \phi_2} \mathbb{E}_{\mathbf{x}_{ij}\sim X_1,X_2}\left[\gamma\left( \mathcal{L}(\mathbf{x}_i,\mathbf{z}_i) +  \mathcal{L}(\mathbf{x}_j,\mathbf{z}_j)\right) + \right. ...\\ ...\lambda \left(\mathcal{L}(\mathbf{z}_i,\mathbf{z}_s) +  \mathcal{L}(\mathbf{z}_j,\mathbf{z}_s)\right) \left. \right]
\end{split} 
\label{eq:full}
\end{equation}

\vspace{-1em}

Here, $\gamma$ and $\lambda$ are hyperparameters that weight the outer and nested VAE ELBO functions respectively. Note that we optimise over all parameters $\phi_1$, $\theta_1$,  $\phi_2$ and $\theta_2$ jointly. In summary, we propose to use VAEs to simultaneously learn both the latent factors $\mathbf{z}_i$ for image $\mathbf{x}_i$ and the latent factors $\mathbf{z}_j$ for image $\mathbf{x}_j$, while ensuring a sufficient and minimal representation $\mathbf{z_s}$ exists between these latent factors. The network architecture is depicted in Figure \ref{fig:nested}. Note that, in practice, we find that feeding the nested VAE the latent codes $\mathbf{\boldsymbol{\mu}}_i$ and $\mathbf{\boldsymbol{\mu}}_j$ rather than  $\mathbf{z}_i$ and $\mathbf{z}_j$ occasionally yields better performance. Furthermore, we also find that the $\beta$ KL weight for the nested VAE should be set close to, or equal to zero for the best results. This is coherent with the application of IB to the derivation of common factors and therefore does not contradict the formulation: $\mathbf{z}_s$ is being derived from the commonality between the parameters $\boldsymbol{\mu}_i$ and $\boldsymbol{\mu}_j$ of the latent random variables $\mathbf{z}_i$ and $\mathbf{z}_j$ respectively, which have already been prior regularized by the outer VAE. We can therefore adjust the IB aspect of NestedVAE shown in Eq. \ref{eq:markov} to:

\vspace{-1em}
\begin{equation}
\boldsymbol{\mu}_i \stackrel{q(z | \mu_i)}{\longrightarrow} \mathbf{z}_s \stackrel{p( \mu_j| z)}{\longrightarrow} \boldsymbol{\mu}_j
\end{equation}
\vspace{-1em}

The full training process algorithm for NestedVAE is shown in the supplementary material.

\section{Prior Work: A Unifying Perspective}
Previous work has aimed to achieve a range of seemingly distinct goals which include disentanglement, domain/attribute invariance, fair encodings and bias reduction, generalization, and common causes. In this section, we review examples of such work, whilst drawing attention to the significant commonality between the goals. By noting the commonality, we hope that progress in one area may be leveraged to make progress in the others.

We have identified the problem of achieving domain invariance, which is to transfer learning between domains whilst being invariant to the confounders and covariates unique to each domain. When such confounders are considered to be `sensitive' attributes, achieving domain invariance may also be considered to be achieving bias reduction, fairness, or demographic parity; when such confounders cause distributional shift, achieving invariance may be considered to be achieving model generalization. Such tasks either require that the confounding information is `forgotten' or ignored, or that it be disentangled from the domain invariant (i.e. task relevant) factors. However, the task of forgetting is often treated as being distinct from disentanglement. We argue that these tasks complement each other: one researcher's disentangled, generative attribute may be another's confounder. For instance, in facial recognition, the identity of an individual should be predicted from an image in such a way that the prediction is invariant to the head-pose and facial expression; it does not benefit the model to provide a different identity representation for a different head-pose. For such an application, a method may either `learn to forget' head-pose, or to disentangle head-pose from identity such that the information encoding identity is independent of, and separable from, the information for pose. In both disentanglement and domain invariance, task-relevant information needs to be separated from task-irrelevant information. 

Furthermore, many of the models utilised for disentanglement are deep latent variable models \cite{higgins, kulkarni, disentanglement, gabbay2019, infovae, kumar2, locatello}. Such models aim to infer the generative or causal factors behind the observed data. As such, using these models to identify factors which are common between domains (as NestedVAE does) becomes equivalent both to identifying the common causes as well as to identifying the factors which generalize across domains. Much of the prior work on unsupervised disentanglement \cite{higgins, sepliarskaia2019, shu2019, gabbay2019, disentanglement, infovae, kumar2, locatello} therefore also indirectly contributes to the field of domain invariance and fairness. Indeed, recent work \cite{locatello2019fairness} has specifically explored the connection between disentanglement and fairness.

Previous research has sought to disentangle and/or learn invariant representations by incorporating supervision with fully supervised VAEs \cite{kulkarni, creager2019}, semi-supervised VAEs \cite{louizos2017, moyer1, DIVA, siddharth2017}, adversarial training \cite{goodfellow2, hadad, ganin1, shankar2018, xie2018, Wang2019, lample2018, mathieu2016, press1, zheng2018, adeli2019}, Shannon Mutual Information regularization \cite{klys2018, MAE} and paired images with auxiliary classifiers \cite{bousmalis2016, JADE}. In other scenarios, we may only have access to indirect supervision for $\mathbf{c}$ \eg in the form of grouped or paired images \cite{szabo2018, feng2018, abid2019} or pairwise similarities \cite{chen2019, chen2019b}. In such cases, previous work has incorporated such weak supervision into VAEs \cite{ruiz2019, denton2017, MLVAE, chen2019, gatedvae}, cycle-consistent networks \cite{jha2018, DRIT}, autoencoders \cite{feng2018}, and autoencoders with adversarial training \cite{szabo1}. In scenarios whereby no supervision is available to assist in learning invariant embeddings, unsupervised approaches are possible which may involve testing for disentanglement and interventional robustness \cite{suter2018, locatello}. Existing methods that aim to achieve domain invariance and/or disentanglement therefore vary in the level of incorporation of supervision. 
 
 Acquiring high quality labelled datasets is both time consuming and expensive, and supervised methods such as those that require labels for class, domain, and/or covariate (\eg as for \cite{adeli2019, DIVA}) may not always be feasibile. Disentanglement may allow for an embedding to be learned such that the undesired covariate is identifiable or extricable at a later time for a specific downstream task. However, the efficacy of completely unsupervised methods for disentanglement has recently been shown to vary as much by random-seed as by architecture and design \cite{locatello}.

Given the disadvantages of both fully supervised and fully unsupervised methods, it is pertinent to consider methods that incorporate minimal levels of weak supervision. Despite some overlap between definitions \cite{goodfellow}, weak supervision is generally used to describe the scenario whereby labels are available but the labels only relate to a limited number of factors \cite{szabo1}. Semi-supervision, in contrast, describes the scenario whereby fully informative labelling is available but only for a subset of the data \cite{kingma3}. Whilst adversarial methods have been shown to work well for `forgetting' information, they are also notoriously difficult and unreliable to train \cite{moyer1, lezama, gabbay2019}. Further, previous work has highlighted that adversarial training is unnecessary, and that non-adversarial training can achieve comparable or better results \cite{moyer1,gabbay2019}. Given the disadvantages of adversarial training and the comparable success of VAEs, we consider developing a new method using the VAE as a foundation. VAEs are a form of latent variable model \cite{kingma} and are therefore suitable for the task of deriving invariant representations from observations with limited supervision.

 The closest prior work to ours in terms of architectural similarity is probably Joint Autoencoders for Disentanglement (JADE) \cite{JADE}. JADE pairs images according to a common label, feeds each image through a separate VAE and uses a partition from each VAE latent space to predict the shared label, thereby attempting to disentangle label relevant information from label irrelevant information. JADE is evaluated according to its capacity for transfer learning from one, data abundant domain (the full MNIST dataset \cite{MNIST}) to a data scarce domain (chosen to be a reduced version SVHN dataset \cite{SVHN}). The NestedVAE differs in that we do not use labels indicating the domain, thereby significantly weakening the level of explicit supervision. Work by \cite{denton2017} pairs images according to whether or not they derive from the same video sequence, and is classified by the researchers as being an unsupervised method. We take a similar approach with NestedVAE by pairing images, but broaden the input pairings beyond those from the same video sequence to those that are from two domains but that share some common attribute(s). The result is a network that `forgets' information specific to each domain, and learns factors common to both without adversarial training, and with only minimal, weak supervision.

\begin{figure*}
\centering
\includegraphics[width=0.75\linewidth]{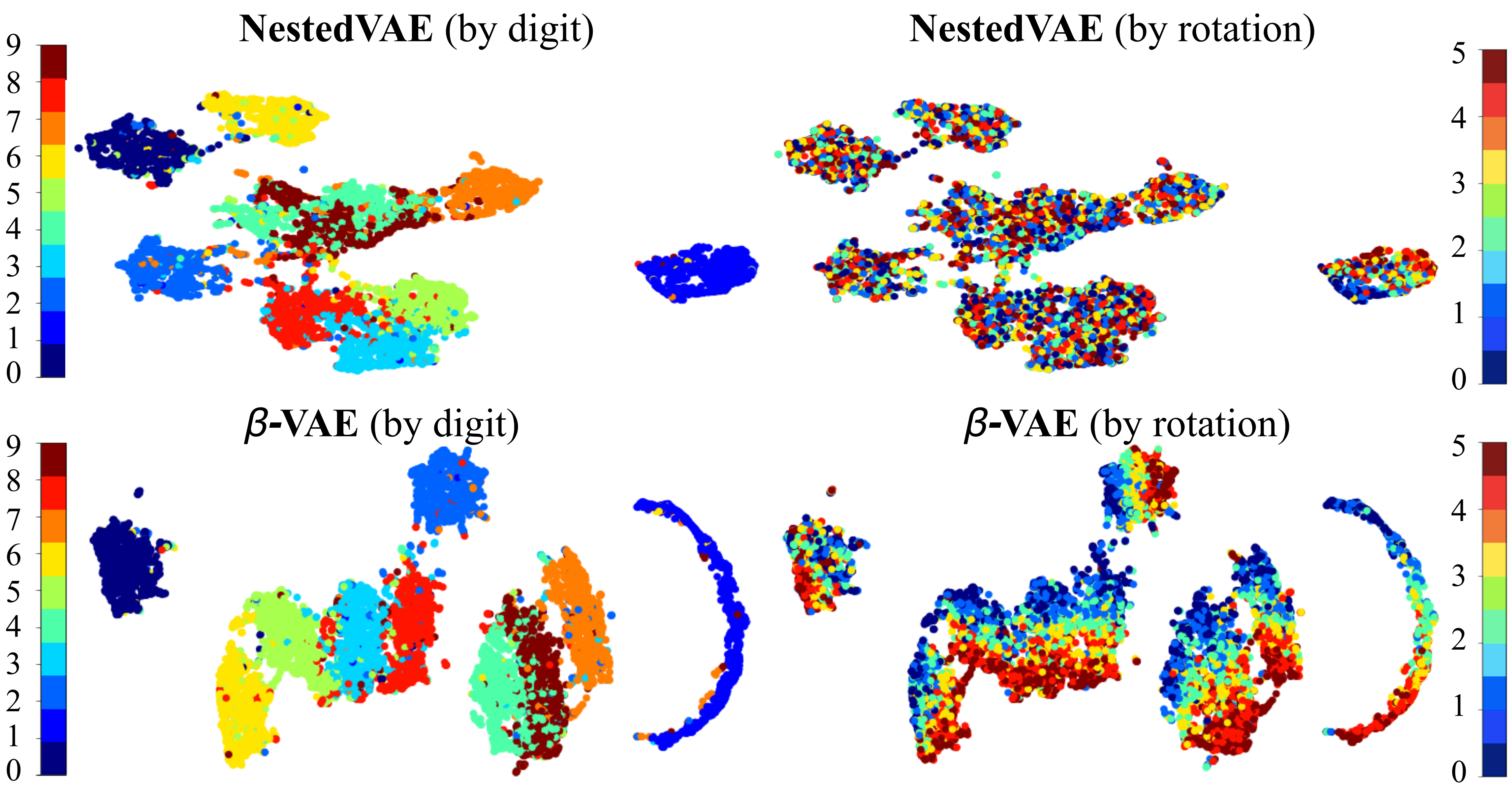}
\caption{UMAP projections of the representations learned from the rotated MNIST dataset by NestedVAE and $\beta$VAE. The representations are coloured according to digit class (left) and rotation domain (right). It can be seen that NestedVAE representations contain significantly less information about rotation than do the $\beta$VAE representations. Best viewed in color.}
\label{fig:rmnist}
\end{figure*}

\begin{table*}[]
\centering
\small
\begin{tabular}{lllllll}
\hline
 & Transfer Domain & \textbf{Nested (ours)}                                      & \textbf{$\boldsymbol{\beta}$VAE}      & \textbf{infoVAE}     & \textbf{DIP-VAE-I}         & \textbf{DIP-VAE-II}        \\ \hline
Digit Classification & $0^\circ$   & \textbf{0.708$\pm$0.211} & 0.551$\pm$0.262 & 0.629$\pm$0.141  & 0.561$\pm$0.213 & 0.519$\pm$0.274 \\
(higher is better)& $15^\circ$  & \textbf{0.696$\pm$0.202} & 0.546$\pm$0.261 & 0.633$\pm$0.132  & 0.597$\pm$0.189 & 0.527$\pm$0.270 \\
& $30^\circ$  & \textbf{0.714$\pm$0.152} & 0.555$\pm$0.251 & 0.657$\pm$0.076  & 0.602$\pm$0.206 & 0.539$\pm$0.244 \\
& $45^\circ$  & \textbf{0.738$\pm$0.124} & 0.575$\pm$0.212 & 0.681$\pm$0.056  & 0.587$\pm$0.208 & 0.510$\pm$0.275 \\
& $60^\circ$  & \textbf{0.721$\pm$0.127} & 0.573$\pm$0.203 & 0.682$\pm$0.057 & 0.577$\pm$0.224 & 0.487$\pm$0.278 \\
& $75^\circ$  & \textbf{0.647$\pm$0.250} & 0.509$\pm$0.249 & 0.588$\pm$0.183  & 0.417$\pm$0.203 & 0.488$\pm$0.253\\

  $\bar{\Delta}_{\mbox{adj}}$ Parity &  n/a & \textbf{0.664} & 0.525 & 0.603 &0.486 & 0.492\\ \hline
Rotation Classification &$0^\circ$   & \textbf{0.373$\pm$0.029} & 0.530$\pm$0.011 & 0.523$\pm$0.005 & 0.511$\pm$0.012 & 0.541$\pm$0.007 \\
(lower is better) &$15^\circ$  & \textbf{0.343$\pm$0.008} & 0.534$\pm$0.005 & 0.516$\pm$0.008 & 0.493$\pm$0.007 & 0.547$\pm$0.005 \\
&$30^\circ$  & \textbf{0.295$\pm$0.050} & 0.534$\pm$0.007 & 0.546$\pm$0.005 & 0.494$\pm$0.005 & 0.538$\pm$0.006 \\
&$45^\circ$  & \textbf{0.316$\pm$0.025} & 0.532$\pm$0.006 & 0.540$\pm$0.001 & 0.493$\pm$0.007 & 0.541$\pm$0.003 \\
&$60^\circ$  & \textbf{0.321$\pm$0.014} & 0.534$\pm$0.005 & 0.542$\pm$0.006 & 0.495$\pm$0.007 & 0.549$\pm$0.007 \\
&$75^\circ$  & \textbf{0.347$\pm$0.057} & 0.517$\pm$0.012 & 0.509$\pm$0.010 & 0.496$\pm$0.016 & 0.518$\pm$0.012 \\
\hline \hline
\end{tabular}
\caption{Average F1-scores and standard errors over 10 runs for digit class (higher is better) and rotation domain (lower is better) classification. NestedVAE is compared against $\beta$VAE \cite{higgins}, infoVAE \cite{infovae}, and DIP-VAE-II \cite{kumar2}. For digit classification, `Transfer domain' refers to the test domain used for classifying the image representations, and this domain is not used during training (\ie domain $0^\circ$ means the network has been trained on domains $15^\circ - 75^\circ $ and is being tested on data from domain $0^\circ$). For rotation classification, the setup is similar in that it represents the domain on training not used during training, although all domains are used for testing (\ie domain $0^\circ$ means the network has been trained on  domains $15^\circ - 75^\circ $ and is being tested on data from ALL domains). We see that NestedVAE learns more informative representations for digit classification than the alternatives, as well as `forgetting' more domain specific information. $\bar{\Delta}_{\mbox{adj}}$ is the average parity metric presented in Eq. \ref{eq:adjustedparity}. Best results are shown in \textbf{bold}. }
\label{tab:rmnist}
\end{table*}

 \section{Evaluation of NestedVAE}
 In light of the overlap between domain/attribute invariance, fairness, and bias reduction discussed in the previous section, we evaluate NestedVAE on a range of tasks. NestedVAE is first evaluated for domain/attribute invariance and change detection on a synthetic dataset with ground-truth factors: rotated MNIST \cite{rotatedMNIST, MNIST}. For this first evaluation, NestedVAE is compared against $\beta$-VAE \cite{higgins} (which increases the pressure on the KL-divergence loss), infoVAE \cite{infovae} (which minimises maximum mean discrepancy) and DIP-VAE-I and DIP-VAE-II \cite{kumar2}. For a non-synthetic evaluation, we test for fairness and bias reduction with biological sex prediction across individuals of different race using the UTKFace dataset \cite{UTKFace}, and compare with $\beta$-VAE and DIP-VAE-I. Additional results can be found in the supplementary material.
 
\subsection{Adjusted Parity Metric}
For evaluation of domain invariance we propose a (to the best of our knowledge) new parity metric that accounts for both discrepancies in accuracy between domains as well as classifier accuracy or normalized regressor performance. The metric is referred to in this work as the adjusted parity metric (adjusted for accuracy) and is defined as follows:

\begin{equation}
\Delta_{\mbox{adj}} = \bar{\hat{S}}(1-2\sigma_{\mbox{acc}})
\label{eq:adjustedparity}
\end{equation}
\vspace{-1em}

Here, $\bar{\hat{S}}$ is the average accuracy\footnote{Alternatively, the F1 score may be used, which is already normalized to fall between [0,1].} of the classifier over the domains, normalized to be between [0,1] according to the baseline accuracy of a random prediction. For example, if we have equal chance of predicting any of the 10 MNIST digits by random chance, the baseline is 0.1. $\sigma_{\mbox{acc}}$ is the standard deviation of the normalized classifier accuracies. Any classifier that is minimally consistent \textit{or} minimally accurate will have $\Delta_{\mbox{adj}} = 0$ and any classifier that is maximally consistent \textit{and} maximally accurate will have $\Delta_{\mbox{adj}} = 1$. This metric was motivated by the fact that although a representation may be domain or attribute invariant, this does not imply that it is also a good classifier: it must also be informative for the intended task.

\subsection{Models}
  For the purposes of the evaluations in this work, the VAEs that constitute NestedVAE do not deviate from the `vanilla' implementations, in that they have isotropic Gaussian priors and approximating distributions \cite{kingma}. The outer VAE $\beta$ KL weight is increased gradually from zero and then annealed during training \cite{sonderby, higgins, disentanglement}. Other more exotic formulations of the VAE may certainly be implemented within the NestedVAE formulation (\eg see \cite{alemi, surgery, tomczak,louizos, infovae, cremer, kumar2}). However, the focus of this work is on the adaptation of the general VAE framework for purposes of domain invariance, rather than the optimality of the VAE itself. Full details of the NestedVAE network architectures used for the experiments can be found in the supplementary material.

\begin{table}[H]
\centering
\small
\begin{tabular}{lll}
\hline
 \textbf{Method} & \textbf{Accuracy} &  \\ \hline
Outlier AutoEncoder\cite{Xia2015} & 0.5427 &  \\
VAE \cite{kingma} & 0.5495 &  \\ 
Clustering AutoEncoder \cite{Aytekin2018} & 0.5514 &  \\
Reconstruction Prob. VAE \cite{An2015} & 0.5724 &  \\
Adversarial VAE \cite{mathieu2016}  & 0.5834 &  \\
Multi-Level VAE \cite{MLVAE} & 0.6072 &  \\
Rare-Event VAE \cite{hamaguchi} & 0.7166 &  \\
NestedVAE (ours) & \textbf{0.7380} & \\ \hline \hline
\end{tabular}
\caption{Change detection accuracy on rotated MNIST. L2 distances between pairs of representations where images are paired according to whether they contain the same (no change) or different (change) digits. K-means clustering is then used to group the representation distances. Alternative results taken from \cite{hamaguchi}. The best result is shown in \textbf{bold}.}
\label{tab:changedetection}
\end{table}
\vspace{-1em}

\subsection{Rotated MNIST}
\label{sec:rmnist}
The rotated MNIST training dataset is generated as follows \cite{rotatedMNIST, DIVA}: for each digit class, 100 random samples are drawn and 6 rotations of $\{0^\circ, 15^\circ, 30^\circ, 45^\circ, 60^\circ, 75^\circ\}$ are applied resulting in $(100\times10\times6) = 6000$ images (one tenth the size of the original MNIST training set). This is repeated to produce a non-overlapping test set of the same structure. For each training pair, a random digit class is chosen and two images are chosen with that digit class across a randomly selected pair of (different) rotations. Each rotation group is treated as a domain to which the learned embedding should be invariant. The network is trained on data from 5 out of 6 of the rotation domains, and tested for digit classification performance on the remaining domain (for which the network has seen no samples from the same distribution during training) using a Random Forest classification algorithm. This is then repeated until the network has been trained and tested on all combinations of domains. If the network achieves domain transfer, we should see a good digit classification performance on the test domain. If the network achieves attribute invariance, we should see poor rotation classification performance across all domains.

NestedVAE is then evaluated for its usefulness at change detection using the same methodology as \cite{hamaguchi}. Images are alternately paired according to shared or not shared digit class. If the pair shares the digit class, a `0' ground-truth label is generated, representing no change. If the pair does not share the same digit class, a `1' label is generated, representing a change. The L2 norm is calculated between the representations of the images in each pair, and a k-means clustering algorithm is trained on the L2 distance and evaluated against the labels. 

Finally, the Uniform Manifold Approximation Projection (UMAP) \cite{mcinnes} algorithm is applied to asses domain invariance visually. UMAP is a more recent, more efficient algorithm for manifold projection than the well-known t-distributed Stochastic Neighbor Embedding (tSNE) \cite{tsne}. The results are compared against the best alternative from the quantitative evaluation.

In terms of model parameter values, for $\beta$-VAE, $\beta = 4$ and is annealed during training (as suggested by \cite{higgins, disentanglement}), for DIP-VAE-I, $\lambda_{od} = 10$ and $\lambda_d = 100$, for DIP-VAE-II, $\lambda_{od} = \lambda_d = 250$, and for InfoVAE $\alpha =0 $ and $\lambda_v=500$ (as suggested by \cite{infovae}) where all $\alpha$, $\lambda_{\_}$ parameters represent a weight on the respective component(s) of the models' objective functions. All models were trained for 100 epochs with an ADAM optimizer with a learning rate of 0.0008 and a batch size of 64. NestedVAE had an inner latent dimensionality of 8, whilst the outer-VAE had a latent dimensionality of 10. The nested and outer VAE weights $\gamma = \delta = 0.5$. All alternative models had a latent dimensionality of 10. 

\textbf{Rotated MNIST Results:}
The results for domain and attribute invariance on rotated MNIST dataset are shown in Table \ref{tab:rmnist}. The results show that NestedVAE is significantly better at learning domain irrelevant information (digit class) as well as being much better at forgetting domain specific information (rotation), than the alternative methods. The Adjusted Parity results are presented in the row labelled $\bar{\Delta}_{\mbox{adj}}$. Note that, because F1 ranges from [0,1], we do not need to normalize F1 according to that of a random prediction before computing the Adjusted Parity. The results for the adjusted parity metric $\bar{\Delta}_{\mbox{adj}}$ demonstrate that NestedVAE outperforms the alternatives.

The UMAP projections are shown in Figure \ref{fig:rmnist}. This figure demonstrates that a 2D projection of the rotated MNIST embeddings may be clearly clustered according to digit labels (left). However, when the projections are coloured according to rotation labels (right), it can be seen that $\beta$-VAE encodes rotation vertically, whilst NestedVAE has, as intended, learned embeddings that are invariant to rotation.

The results for the change detection task are shown in Table \ref{tab:changedetection}. Nested VAE is evaluated against 7 other methods, some of which are specifically designed for change detection and which utilise significantly more powerful network architectures than ours \cite{hamaguchi, MLVAE, mathieu2016}. It can be seen that NestedVAE outperforms the best alternative.

\subsection{UTKFace}
The UTKFace dataset \cite{UTKFace} comprises +20k images with labels for race (White, Black, Asian, Indian, or other) and sex (male or female). Previous work has noted the bias in gender prediction software \cite{gendershades, kortylewski2019, merler2019}, particularly in relation to the accuracy of gender prediction for white individuals compared to the (significantly lower) accuracy for black individuals. We note a distinction between biological sex and gender, and assume that any labels in UTKFace are actually for biological sex. This is because, despite UTKFace referring to gender, the actual labels are for `male' and `female' which are terms more sensitively attributed to sex (see \cite{cao2019} for a discussion on the sociological aspects of gender).

\begin{table}[H]
\centering
\small 
\begin{tabular}{lll} \hline
     \textbf{Method}            & $\bar{\Delta}_{\mbox{adj}}$ Parity (Female) & $\bar{\Delta}_{\mbox{adj}}$ Parity (Male) \\ \hline
$\beta$-VAE      & 0.410                & 0.537              \\
DIPVAE-I         & 0.394                & 0.547              \\
NestedVAE (ours) & \textbf{0.641}                & \textbf{0.699  } \\ \hline \hline           
\end{tabular}
\caption{This table shows the Adjusted Parity calculated from F1 scores across race using the UTKFace dataset. Methods with high Adjusted Parity are methods which have high F1 score for the prediction of biological sex and which are consistent across race. Best results are shown in \textbf{bold}. NestedVAE outperforms alternatives.}
\label{tab:UTKFaceparity}
\end{table}

The dataset is first restricted to comprise only white and black individuals. This is done in order to reduce the ambiguity associated with the definition of race across ethnicity, as applied in UTKFace, which uses labels such as `Indian' or 'Other'. Next, the dataset is split into train and test sets, and the training set is further reduced in size such that the number of white individuals is equal to the number of black individuals. We then create 5 versions of the training dataset, whereby the proportion of white individuals is increased from 50\% to 100\%. The model is trained on each of these versions and embeddings for the test set are generated by passing the test set images through the trained model. Gradient boosting classifiers are used to predict sex across white and black individuals and we present the corresponding F1 classification scores and Area Under Receiver Operator Characteristic (AU-ROC) scores.

In terms of model parameter values, for $\beta$-VAE, $\beta = 4$ and is annealed during training (as suggested by \cite{higgins, disentanglement}), for DIP-VAE-I, $\lambda_{od} = 10$ and $\lambda_d = 100$. The models were trained for 1000 epochs with an ADAM optimizer with a learning rate of 0.001 and a batch size of 64.  NestedVAE had an inner latent dimensionality of 50, whilst the outer-VAE had a latent dimensionality of 256. The nested and outer VAE weights $\gamma = \delta = 0.5$. All alternative models had a latent dimensionality of 50. A hyperparameter search yielded gradient boosting classifier parameters as follows: maximum features=50; maximum depth=5; learning rate=0.25, number of estimators=300, minimum samples per split=0.7. Averages and standard deviations are acquired over 5 runs.

\begin{figure}[!ht]
\centering
\includegraphics[width=1\linewidth]{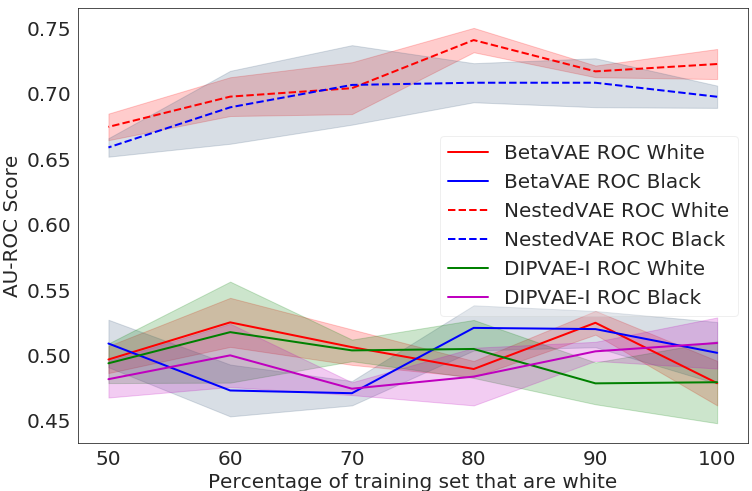}
\caption{Area Under Receiver Operator Characteristic (AU-ROC) scores for models trained on datasets with varying proportions of white and black individuals. NestedVAE significantly outperforms alternatives. Best viewed in color.}
\label{fig:UTKFace}
\end{figure}

\textbf{UTKFace Results:}
The results for Adjusted Parity are shown in Table \ref{tab:UTKFaceparity}. These results provide a measure of consistency and performance (F1 score) of the classifier for the prediction of biological sex across race domains. It can be seen that NestedVAE outperforms alternatives, and also shows the smallest discrepancy in Adjusted Parity between female and male classification performance (0.699 for male, compared with 0.641 for female). Notably, sex is poorly predicted using embeddings from the other models. The poor prediction could be because the alternative models have embedded sex as a continuous variable (e.g. degrees of masculinity/femininity) which is entangled with other appearance dimensions, whereas NestedVAE has been explicitly trained using binary pairings of sex, thereby providing significant inductive bias. The results for the Area Under Receiver Operator Characteristic (AU-ROC) score are shown in Figure \ref{fig:UTKFace}. These results demonstrate the classifier performance for predicting biological sex for black individuals and white individuals using embeddings from models trained on data varying in the proportion of white and black individuals. Interestingly, we do not see a large variation across the training sets, suggesting that the information about sex encoded in the network embeddings is not substantially confounded by race. Nevertheless, NestedVAE clearly outperforms the alternative methods by a significant margin in its ability to isolate the common factors (i.e. factors relating to sex).

\section{Conclusion and Further Work}
 NestedVAE provides a means to learn representations that are invariant to the covariates specific to domains, whilst being able to isolate the common causes across domains. The method combines the theory of deep latent variable VAE models with Information Bottleneck principle and is trained on pairs of images with common factors and where the two images in a pair are sampled from different domains. Results demonstrate NestedVAE's superior performance for achieving domain invariance, change detection, and sex prediction. We have also presented a new (to the best of our knowledge) `adjusted parity metric' in order to facilitate comparison between methods with significantly different classification performance.
 
The principles behind NestedVAE can be applied to more exotic VAEs, and even non-VAEs. Further work should explore the application of the principles to different models.

%-------------------------------------------------------------------------

%-------------------------------------------------------------------------

%-------------------------------------------------------------------------

% {\small
% \bibliographystyle{ieee_fullname}
% \bibliography{egbib}
% }

\clearpage

\section{Supplementary Material}

\subsection{Overview of Supplementary Material}
\begin{figure*}[]
\centering
\includegraphics[width=0.9\linewidth]{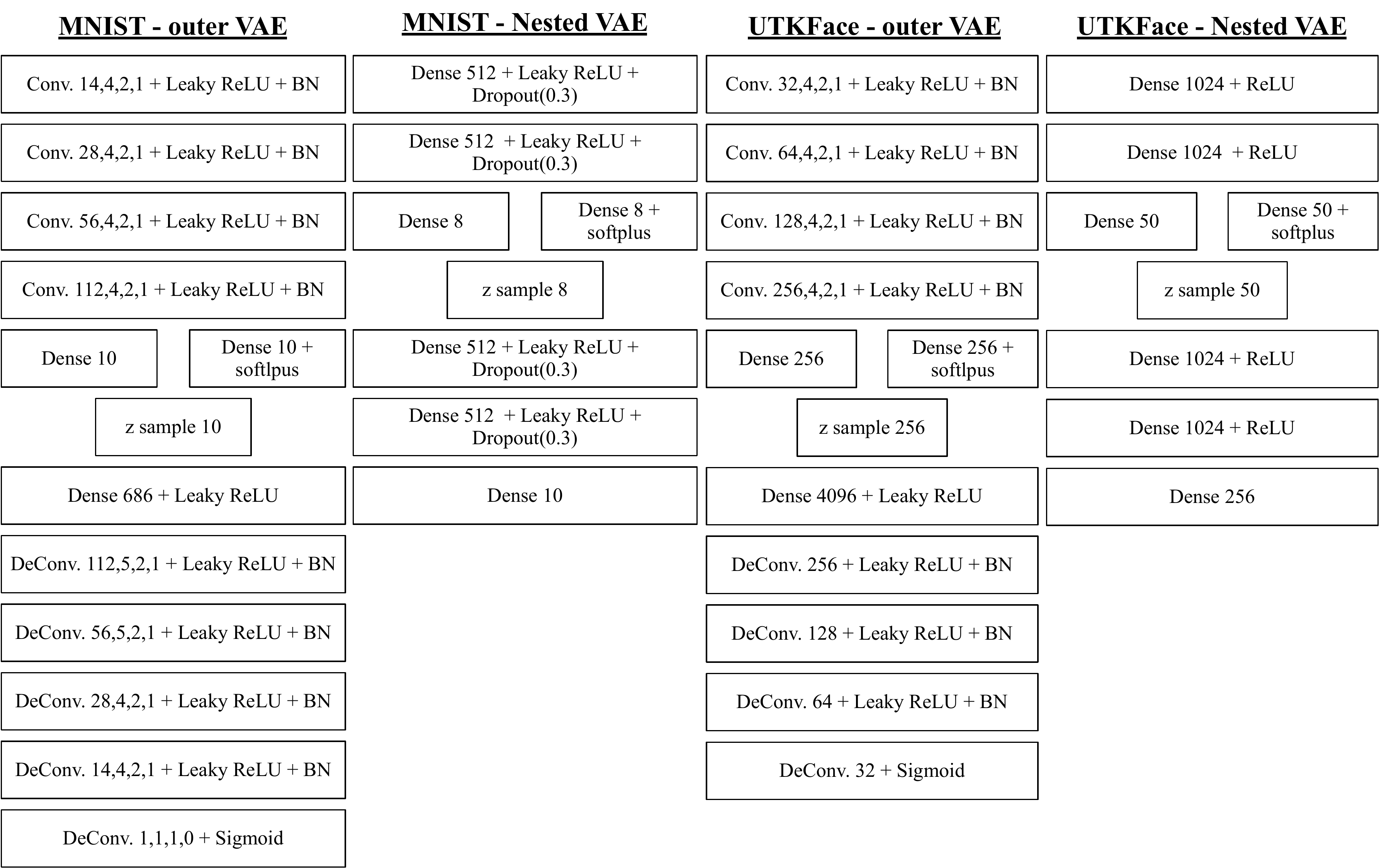}
\caption{Network architectures used for the reported experiments on rotated MNIST and UTKFace datasets.}
\label{fig:mnistdsprites}
\end{figure*}

 This material supplements the paper `NestedVAE: Isolating common factors via weak supervision.' Firstly, we present the algorithm for training NestedVAE in Algorithm 1. Secondly, we present additional results for biological sex prediction on the UTKFace dataset. Finally, we present architectures used for the MNIST and UTKFace experiments in Figure \ref{fig:mnistdsprites}.

 \begin{algorithm}
 \caption{\textsc{NestedVAE} training procedure.}
 \begin{algorithmic}[1]
 \renewcommand{\algorithmicrequire}{\textbf{Input:}}
 \renewcommand{\algorithmicensure}{\textbf{Output:}}
 \REQUIRE Image pairs $\mathbf{x}_i \sim X_1$ and $\mathbf{x}_j \sim X_2$ $i,j \in [1,N]$, learning rate $\alpha$, loss weights $\gamma$, $\lambda$, 
 \ENSURE  NestedVAE $\{\theta_1, \phi_1, \theta_2, \phi_2\}$
 \\ \textit{Initialisation} : 
  \STATE Random init. \cite{glorot3} $\longrightarrow \{\theta_1, \phi_1, \theta_2, \phi_2\}$ 
 \\ \textit{Training}
   \FOR {$s = 0: N/\mbox{batch size}$} 
  \STATE Sample batch $\mathbf{x}_{i,j}$
  \\ \textit{Outer VAE computations} : 
  \STATE Compute batch $\{\boldsymbol{\mu}_{i,j}, \boldsymbol{\sigma}_{i,j}\}= \mbox{Enc}( \mathbf{x}_{i,j})$
  \STATE Sample batch $\mathbf{z}_{i,j}\sim N(\boldsymbol{\mu}_{i,j},\boldsymbol{\sigma}_{i,j}) $
  \STATE Compute batch $\mathbf{\hat{x}}_{i,j} = \mbox{Dec}(\mathbf{z}_{i,j})$
  \STATE Compute batch loss MSE$(\mathbf{\hat{x}}_{i,j}, \mathbf{x}_{i,j})$
    \STATE Compute batch loss $\beta$ KL$\left[(\boldsymbol{\mu}_{i,j}, \boldsymbol{\sigma}_{i,j}), N(0,I) \right]$
    \\ \textit{Nested VAE computations} : 
    \STATE Compute batch $\{\boldsymbol{\mu}_{Nest-i}, \boldsymbol{\sigma}_{Nest-i}\}= \mbox{Enc}( \boldsymbol{\mu}_{i})$
    \STATE Sample batch $\mathbf{z}_{s}\sim N(\boldsymbol{\mu}_{Nest-i},\boldsymbol{\sigma}_{Nest-i}) $
    \STATE Compute batch $\boldsymbol{\hat{\mu}}_{i} = \mbox{Dec}(\mathbf{z}_{s})$
   \STATE Compute batch loss MSE$(\boldsymbol{\hat{\mu}}_{i},\boldsymbol{{\mu}}_j) $
   \STATE Compute batch loss \newline $\beta_{Nest}$KL$\left[(\boldsymbol{\hat{\mu}}_{Nest-i},\boldsymbol{{\sigma}}_{Nest-i}), N(0,I)\right] $
   \\ \textit{Combine} : 
   \STATE Weight VAE and NestedVAE losses by $\gamma$, $\lambda$ respectively and sum.
   \STATE Backpropagate gradients and update w/ ADAM \cite{adam} and learning rate $ =\alpha$

  \ENDFOR
 \end{algorithmic} 
\end{algorithm}
 
%%%%%%%%%%%%%%%%%%%%%%%%%%%%%%%%%%%%%%%%%%%%%%%%%%
%%     DSPRITES %%%%%%%%%%%%%%%%%%%%%%%%%%%%%%%%%%
%%%%%%%%%%%%%%%%%%%%%%%%%%%%%%%%%%%%%%%%%%%%%%%%%%

\subsection{UTKFace}
\begin{figure}[H]
\centering
\includegraphics[width=1\linewidth]{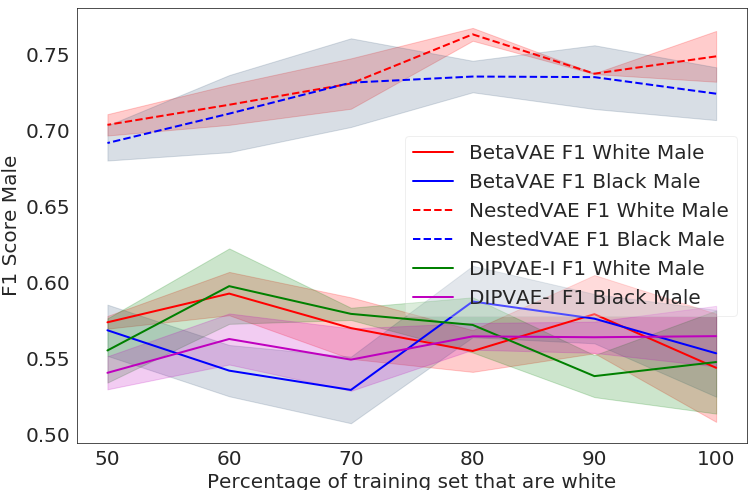}
\caption{F1 scores for prediction of male sex using embeddings from models trained on datasets with varying proportions of white and black individuals. Best viewed in color.}
\label{fig:UTKFacemale}
\end{figure}

\begin{figure}[H]
\centering
\includegraphics[width=1\linewidth]{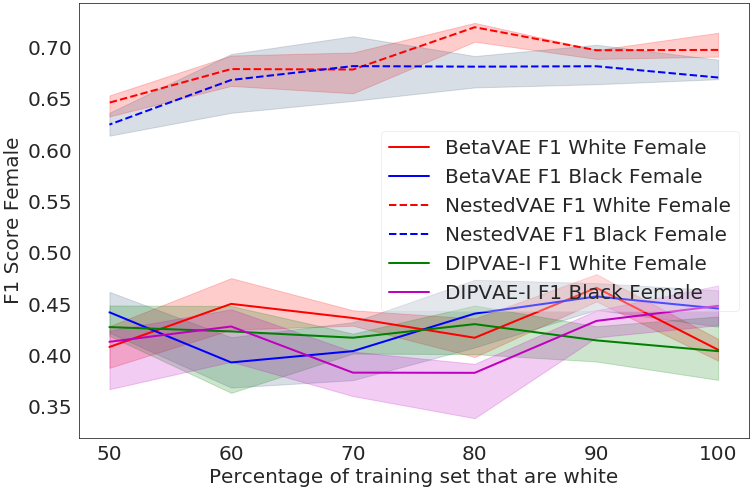}
\caption{F1 scores for prediction of female sex using embeddings from models trained on datasets with varying proportions of white and black individuals.}
\label{fig:UTKFacefemale}
\end{figure}

The UTKFace dataset \cite{UTKFace} evaluation procedure is described in the main paper. The results for F1-score for males across race and females across race are shown in Figures \ref{fig:UTKFacemale} and \ref{fig:UTKFacefemale} respectively. The complete F1-scores are shown in Table \ref{tab:utkface}. It can be seen that NestedVAE significantly outperforms alternatives in all cases, and extracts the common factors for sex.

\vspace{-1em}

\begin{table*}
\small
\centering
\begin{tabular}{llllll} \hline
                 & \textbf{Percent White} & \textbf{Black Female}    & \textbf{Black Male}      & \textbf{White Female}    & \textbf{White Male}       \\ \hline
$\beta$-\textbf{VAE}      & 100           & 0.446$\pm$0.018 & 0.553$\pm$0.029 & 0.405$\pm$0.011 & 0.544$\pm$0.036             \\
                 & 90            & 0.457$\pm$0.013 & 0.576$\pm$0.016 & 0.466$\pm$0.013 & 0.579$\pm$0.026            \\
                 & 80            & 0.441$\pm$0.033 & 0.587$\pm$0.023 & 0.417$\pm$0.019 & 0.555$\pm$0.014             \\
                 & 70            & 0.404$\pm$0.028 & 0.529$\pm$0.022 & 0.436$\pm$0.008 & 0.570$\pm$0.020              \\
                 & 60            & 0.393$\pm$0.025 & 0.542$\pm$0.017 & 0.450$\pm$0.025  & 0.593$\pm$0.014              \\
                 & 50            & 0.442$\pm$0.020 & 0.568$\pm$0.017 & 0.408$\pm$0.020 & 0.574$\pm$0.004              \\
\textbf{DIPVAE-I}         & 100           & 0.448$\pm$0.019 & 0.565$\pm$0.020 & 0.404$\pm$0.034 & 0.548$\pm$0.020             \\
                 & 90            & 0.434$\pm$0.016 & 0.564$\pm$0.010 & 0.415$\pm$0.014 & 0.538$\pm$0.010             \\
                 & 80            & 0.383$\pm$0.044 & 0.564$\pm$0.009 & 0.430$\pm$0.018 & 0.572$\pm$0.009              \\
                 & 70            & 0.383$\pm$0.023 & 0.549$\pm$0.021 & 0.417$\pm$0.004 & 0.579$\pm$0.021             \\
                 & 60            & 0.428$\pm$0.034 & 0.564$\pm$0.017 & 0.423$\pm$0.025 & 0.597$\pm$0.017              \\
                 & 50            & 0.413$\pm$0.046 & 0.540$\pm$0.011 & 0.428$\pm$0.021 & 0.555$\pm$0.011              \\
\textbf{NestedVAE} (ours) & 100           & \textbf{0.671$\pm$0.002} & \textbf{0.724$\pm$0.017} & \textbf{0.698$\pm$0.007} & \textbf{0.749$\pm$0.017}              \\
                 & 90            & \textbf{0.682$\pm$0.018} & \textbf{0.735$\pm$0.021} & \textbf{0.698$\pm$0.008} & \textbf{0.737$\pm$0.001}              \\
                 & 80            & \textbf{0.682$\pm$0.020} & \textbf{0.735$\pm$0.010} & \textbf{0.720$\pm$0.014} & \textbf{0.763$\pm$0.004}              \\
                 & 70            & \textbf{0.682$\pm$0.034} & \textbf{0.731$\pm$0.029} & \textbf{0.679$\pm$0.023} & \textbf{0.731$\pm$0.017}              \\
                 & 60            & \textbf{0.669$\pm$0.032} & \textbf{0.711$\pm$0.025} & \textbf{0.679$\pm$0.016} & \textbf{0.717$\pm$0.013}              \\
                 & 50            & \textbf{0.625$\pm$0.011} & \textbf{0.692$\pm$0.012} & \textbf{0.646$\pm$0.013} & \textbf{0.703$\pm$0.007}            \\ \hline \hline
\end{tabular}
\caption{F1 score results for classifier performance on black females, black males, white females, and white males using embeddings from models trained on data varying in the proportion of white individuals. Results demonstrate superior classification performance with NestedVAE. Best results are shown in \textbf{bold}.}
\label{tab:utkface}
\end{table*}

\end{document}